\documentclass[acmsmall]{acmart}

\AtBeginDocument{%
  }

\setcopyright{acmlicensed}
\copyrightyear{2024}
\acmYear{2024}
\acmDOI{XXXXXXX.XXXXXXX}

\acmJournal{JACM}
\acmVolume{37}
\acmNumber{4}
\acmArticle{111}
\acmMonth{8}

\usepackage{multirow}
\usepackage{enumitem}
\usepackage{xspace}
\usepackage{indentfirst}

\usepackage{multirow}
\usepackage[normalem]{ulem}
\useunder{\uline}{\ul}{}

\usepackage{color}
\definecolor{mygreen}{RGB}{112,173,71} 
\definecolor{myblue}{RGB}{0,112,192} 
\definecolor{myorange}{RGB}{255,141,65}

\newcommand{\secref}[1]{Sec.~\ref{#1}}
\newcommand{\figref}[1]{Fig.~\ref{#1}}

\newcommand{\equref}[1]{Equ.~\ref{#1}}

\newcommand{\modelname}{X-Dreamer\xspace}
\newcommand{\moduleOneBig}{Camera-Guided Low-Rank Adaptation\xspace}

\newcommand{\moduleOneShort}{CG-LoRA\xspace}
\newcommand{\moduleTwoBig}{Attention-Mask Alignment Loss\xspace}
\newcommand{\moduleTwoSL}{Attention-Mask Alignment (AMA) Loss\xspace}

\newcommand{\moduleTwoShort}{AMA loss\xspace}
\newcommand{\dmtet}{\textsc{DMTet}\xspace}
\begin{document}

\title{Creating High-quality 3D Content by Bridging the Gap Between Text-to-2D and Text-to-3D Generation}

\author{Yiwei Ma}
\authornote{Both authors contributed equally to this research.}
\affiliation{%
  \institution{Xiamen University}
  \city{Xiamen}
  \country{China}}
\email{yiweima@stu.xmu.edu.cn}

\author{Yijun Fan}
\authornotemark[1]
\affiliation{%
  \institution{Xiamen University}
  \city{Xiamen}
  \country{China}}
\email{fyj08092000@163.com}

\author{Jiayi Ji}
\affiliation{%
  \institution{Xiamen University}
  \city{Xiamen}
  \country{China}}
\email{jjyxmu@gmail.com}

\author{Haowei Wang}
\affiliation{%
  \institution{Xiamen University}
  \city{Xiamen}
  \country{China}}
\email{wanghaowei@stu.xmu.edu.cn}

\author{Haibing Yin}
\affiliation{%
  \institution{Hangzhou Dianzi University}
  \city{Hangzhou}
  \country{China}}
\email{yhb@hdu.edu.cn}

\author{Xiaoshuai Sun}
\authornote{Corresponding author}
\affiliation{%
  \institution{Xiamen University}
  \city{Xiamen}
  \country{China}}
\email{xssun@xmu.edu.cn}

\author{Rongrong Ji}
\affiliation{%
  \institution{Xiamen University}
  \city{Xiamen}
  \country{China}}
\email{rrji@xmu.edu.cn}

\renewcommand{\shortauthors}{Yiwei Ma et al.}

\begin{abstract}
  In recent times, automatic text-to-3D content creation has made significant progress, driven by the development of pretrained 2D diffusion models. 
    Existing text-to-3D methods typically optimize the 3D representation to ensure that the rendered image aligns well with the given text, as evaluated by the pretrained 2D diffusion model. 
    Nevertheless, a substantial domain gap exists between 2D images and 3D assets, primarily attributed to variations in camera-related attributes and the exclusive presence of foreground objects. Consequently, employing 2D diffusion models directly for optimizing 3D representations may lead to suboptimal outcomes.
    To address this issue, we present \modelname, a novel approach for high-quality text-to-3D content creation that effectively bridges the gap between text-to-2D and text-to-3D synthesis.
    The key components of \modelname are two innovative designs: \moduleOneBig (\moduleOneShort) and \moduleTwoSL.
    \moduleOneShort dynamically incorporates camera information into the pretrained diffusion models by employing camera-dependent generation for trainable parameters. This integration makes the 2D diffusion model camera-sensitive.
    \moduleTwoShort guides the attention map of the pretrained diffusion model using the binary mask of the 3D object, prioritizing the creation of the foreground object. This module ensures that the model focuses on generating accurate and detailed foreground objects.
    Extensive evaluations demonstrate the effectiveness of our proposed method compared to existing text-to-3D approaches. 
    Our project webpage: \textbf{{\url{https://anonymous-11111.github.io/}}}. Our code is available at \textbf{{\url{https://github.com/xmu-xiaoma666/X-Dreamer}}}.
\end{abstract}

\begin{CCSXML}
<ccs2012>
   <concept>
       <concept_id>10002951.10003317.10003371.10003386.10003388</concept_id>
       <concept_desc>Information systems~Video search</concept_desc>
       <concept_significance>500</concept_significance>
       </concept>
   <concept>
       <concept_id>10002951.10003317.10003338.10010403</concept_id>
       <concept_desc>Information systems~Novelty in information retrieval</concept_desc>
       <concept_significance>500</concept_significance>
       </concept>
 </ccs2012>
\end{CCSXML}

\ccsdesc[500]{Applied computing~Media arts}
\ccsdesc[500]{Applied computing~Computer-aided design}
\ccsdesc[500]{Applied computing~Computer-Fine arts}

\keywords{Text-to-3D Generation, Vision and Language, Domain Gap}

% \received[accepted]{26 July 2024}

\maketitle

\section{Introduction}
\label{sec:intro}

\begin{figure}
  \centering
  \includegraphics[width=0.6\columnwidth]{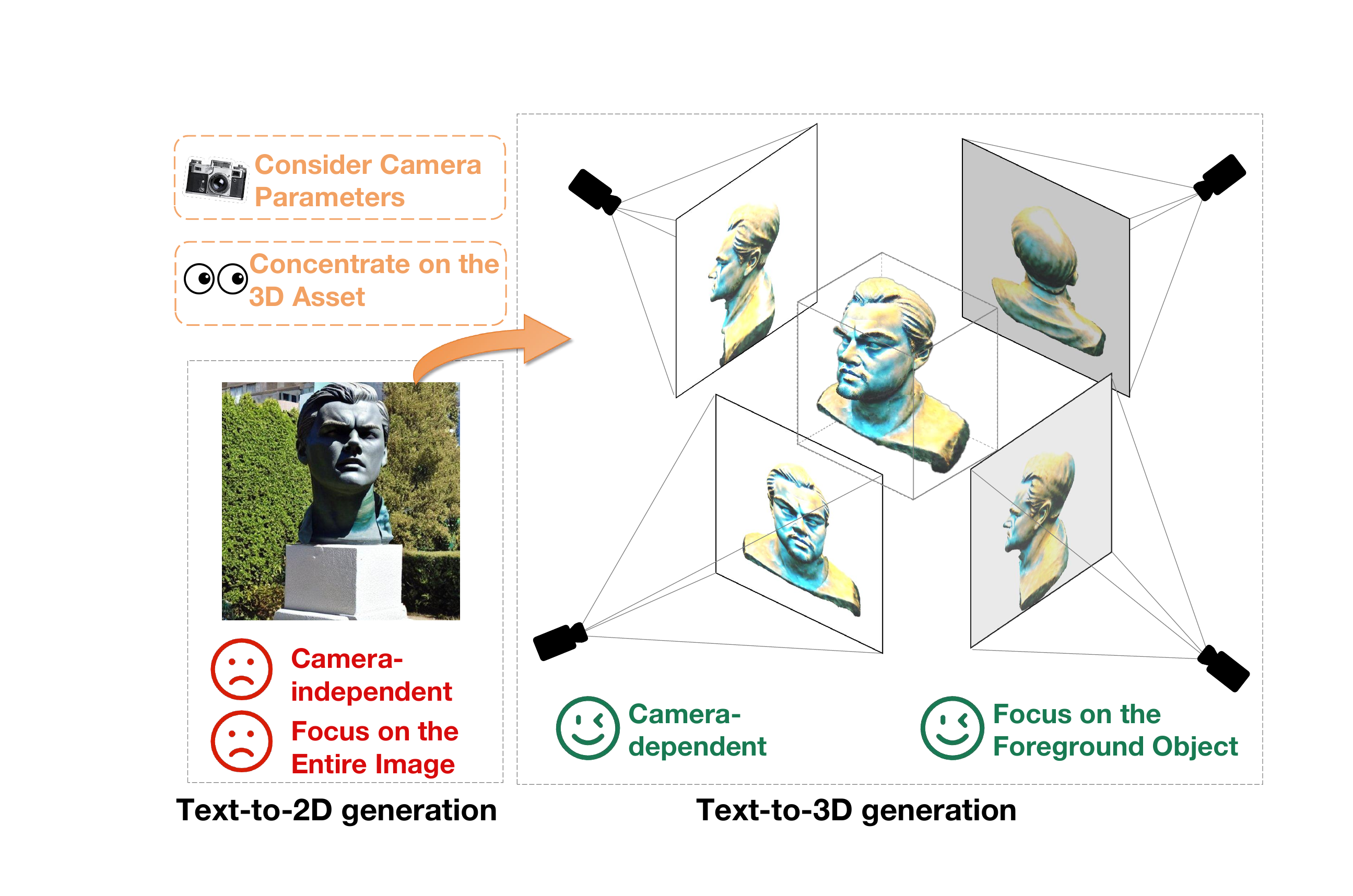}
  %\vspace{-1.5em}
  \caption{The outputs of the text-to-2D generation model (left) and the text-to-3D generation model (right) under the same text prompt, \emph{i.e.,} ``A statue of Leonardo DiCaprio's head''.}
  \label{fig:motivation}
  %\vspace{-1em}
\end{figure}

\begin{figure*}
  \centering
  \includegraphics[width=1.0\columnwidth]{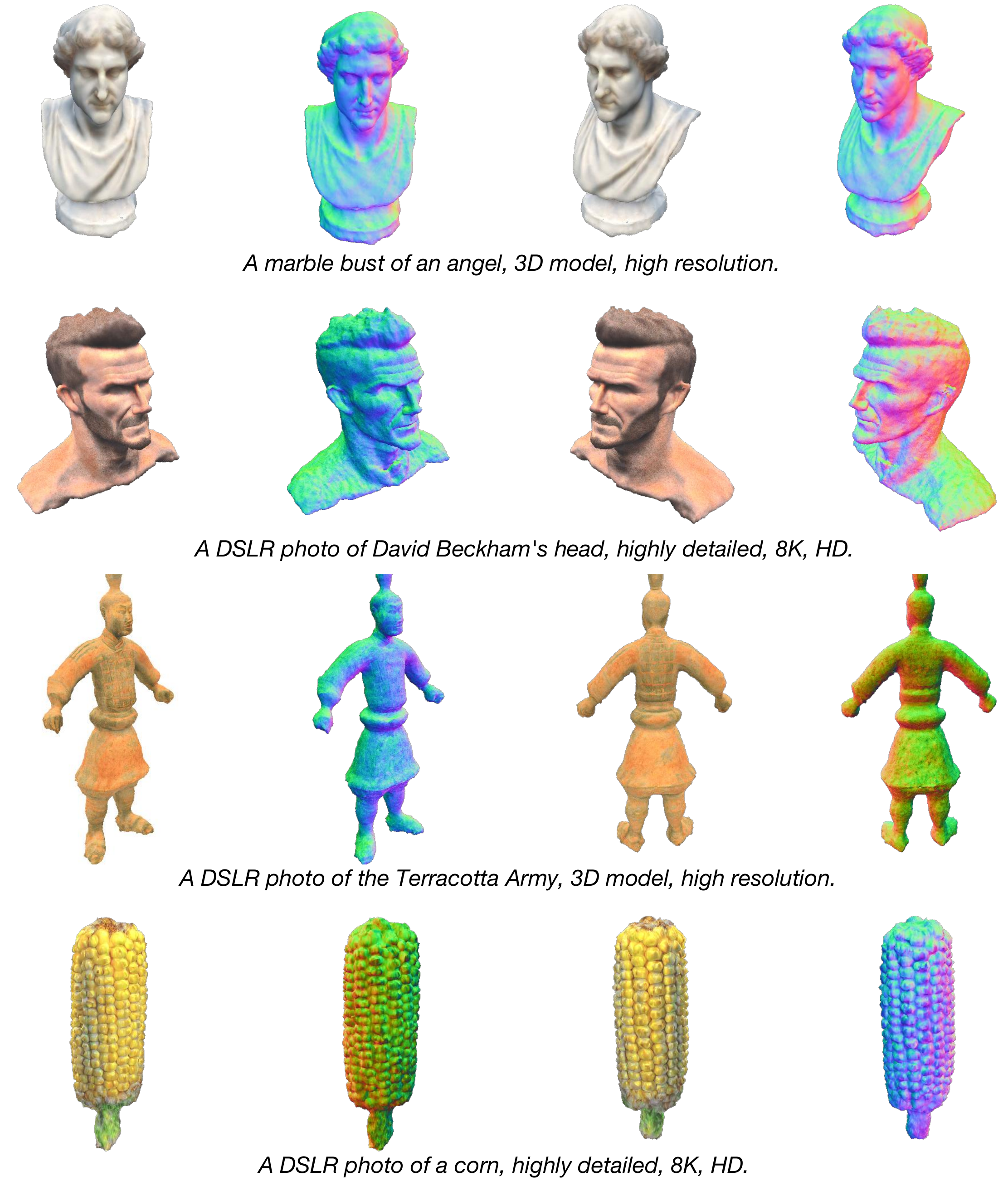}
  \caption{Text-to-3D generation results of the proposed \modelname.}
  \label{fig:morevis1}
\end{figure*}

% 介绍text-to-3D任务
The field of text-to-3D synthesis, which seeks to generate superior 3D content predicated on input textual descriptions, has shown significant potential to impact a diverse range of applications. 
These applications extend beyond traditional areas such as architecture, animation, and gaming, and encompass contemporary domains like virtual and augmented reality.

% 介绍现有的方法
In recent years, extensive research~\cite{chen2023control3d,yu2023points,huang2023avatarfusion} has demonstrated significant performance improvement in the text-to-2D generation task~\cite{ramesh2022hierarchical,saharia2022photorealistic,rombach2022high,balaji2022ediffi,wu24next} by leveraging pretrained diffusion models~\cite{sohl2015deep,ho2020denoising,song2020score} on a large-scale text-image dataset~\cite{schuhmann2022laion}.
Building on these advancements, DreamFusion~\cite{poole2022dreamfusion} introduces an effective approach that utilizes a pretrained 2D diffusion model~\cite{saharia2022photorealistic} to autonomously generate 3D assets from text, eliminating the need for a dedicated 3D asset dataset.
A key innovation introduced by DreamFusion is the Score Distillation Sampling (SDS) algorithm.
This algorithm aims to optimize a single 3D representation, such as NeRF~\cite{mildenhall2021nerf}, to ensure that rendered images from any camera perspective maintain a high likelihood with the given text, as evaluated by the pretrained 2D diffusion model.
Inspired by the groundbreaking SDS algorithm, several recent works~\cite{lin2023magic3d,chen2023fantasia3d,metzer2023latent,wang2023score,wang2023prolificdreamer} have emerged, envisioning the text-to-3D generation task through the application of pretrained 2D diffusion models.

% 引出问题
While text-to-3D generation has made significant strides through the utilization of pretrained text-to-2D diffusion models~\cite{saharia2022photorealistic,zhang2023adding,kumari2023multi}, it is crucial to recognize and address the persistent and substantial domain gap that remains between text-to-2D and text-to-3D generation. 
This distinction is clearly illustrated in \figref{fig:motivation}.
% 
% This gap presents a formidable challenge that directly impacts the optimal performance of utilizing pretrained 2D diffusion models for 3D asset creation.
% 
To begin with, the text-to-2D model produces camera-independent generation results, focusing on generating high-quality images from specific angles while disregarding other angles.
In contrast, 3D content creation is intricately tied to camera parameters such as position, shooting angle, and field of view. 
As a result, a text-to-3D model must generate high-quality results across all possible camera parameters.
This fundamental difference emphasizes the necessity for innovative approaches that enable the pretrained diffusion model to consider camera parameters.
Furthermore, a text-to-2D generation model~\cite{sun2024create,chen2023controlstyle,ruan2023mm,qu2023layoutllm,lu2023painterly,fei2024dysen,fei2024vitron} must simultaneously generate both foreground and background elements while maintaining the overall coherence of the image.
Conversely, a text-to-3D generation model~\cite{raj2023dreambooth3d,xu2023dream3d,lorraine2023att3d} only needs to concentrate on creating the foreground object.
This distinction allows text-to-3D models to allocate more resources and attention to precisely represent and generate the foreground object.
Consequently, the domain gap between text-to-2D and text-to-3D generation poses a significant performance obstacle when directly employing pretrained 2D diffusion models for 3D asset creation.
% 
% The inherent limitations of text-to-2D models in capturing the full scope of 3D scenes, coupled with their emphasis on image quality from specific viewpoints, inevitably lead to suboptimal results when applied to text-to-3D synthesis tasks.

% 介绍我们的方法
In this study, we present a pioneering framework, \modelname, designed to address the domain gap between text-to-2D and text-to-3D generation, thereby facilitating the creation of high-quality text-to-3D content.
Our framework incorporates two innovative designs that are specifically tailored to address the aforementioned challenges.
Firstly, existing approaches~\cite{lin2023magic3d,chen2023fantasia3d,metzer2023latent,wang2023score,ma2022xclip,ma2023towards} commonly employ 2D pretrained diffusion models~\cite{saharia2022photorealistic,rombach2022high} for text-to-3D generation, which lack inherent linkage to camera parameters.
To address this limitation and ensure that our text-to-3D model produces results that are directly influenced by camera parameters, we introduce \emph{\moduleOneBig (\moduleOneShort)} to fine-tune the pretrained 2D diffusion model.
Notably, the parameters of \moduleOneShort are dynamically generated based on the camera information during each iteration, establishing a robust relationship between the text-to-3D model and camera parameters. 
Furthermore, pretrained text-to-2D diffusion models allocate attention to both foreground and background generation, whereas the creation of 3D assets necessitates a stronger focus on accurately generating foreground objects. 
To address this requirement, we introduce \emph{\moduleTwoSL}, which leverages the rendered binary mask of the 3D object to guide the attention map of the pretrained 2D stable diffusion model~\cite{rombach2022high}.
By incorporating this module, \modelname prioritizes the generation of foreground objects, resulting in a significant enhancement of the overall quality of the generated 3D content.
% 
% Through the integration of these two innovative designs, \modelname surpasses the limitations of existing approaches, facilitating the creation of high-quality text-to-3D content that is both camera-dependent and foreground-focused.

We present a compelling demonstration of the effectiveness of \modelname in synthesizing high-quality 3D assets based on textual cues, as shown in \figref{fig:morevis1}.
By incorporating \moduleOneShort and \moduleTwoShort to address the domain gap between text-to-2D and text-to-3D generation, our proposed framework exhibits substantial advancements over prior methods in text-to-3D generation.
In summary, our study contributes to the field in three key aspects:

% \begin{itemize}[itemsep=8pt,topsep=5pt,parsep=0pt,leftmargin=10pt]
% \begin{itemize}[itemsep=2pt,topsep=5pt,parsep=0pt]
\begin{itemize}[]

\item We propose a novel method, \modelname, for high-quality text-to-3D content creation, effectively bridging the domain gap between text-to-2D and text-to-3D generation.  

\item To enhance the alignment between the generated results and the camera perspective, we propose \moduleOneShort, which leverages camera information to dynamically generate \moduleOneShort parameters for 2D diffusion models.

\item To prioritize the creation of foreground objects in the text-to-3D model, we introduce \moduleTwoShort, which utilizes binary masks of the foreground 3D object to guide the attention maps of the 2D diffusion model.

\end{itemize}

\section{Related Work}
\label{sec:relatedwork}

\subsection{Text-to-3D Content Creation}

% 参考Fantasia3D
In recent years, there has been a significant surge in interest surrounding the evolution of text-to-3D generation~\cite{michel2022text2mesh,poole2022dreamfusion,li2023sweetdreamer,chen2023text,li2024focaldreamer,sella2023vox,wu24next,yang20233dstyle,fei2024video,babu2023hyperfields,fei2024enhancing}.
This growing field has been propelled, in part, by advancements in pretrained vision-and-language models, such as CLIP~\cite{radford2021learning}, as well as diffusion models like Stable Diffusion~\cite{rombach2022high} and Imagen~\cite{saharia2022photorealistic}.
Contemporary text-to-3D models can generally be classified into two distinct categories: \emph{the CLIP-based text-to-3D approach} and \emph{the diffusion-based text-to-3D approach}.
The CLIP-based text-to-3D approach~\cite{michel2022text2mesh,ma2023x,lei2022tango,jain2022zero,mohammad2022clip,wei2023taps3d} employs CLIP encoders~\cite{radford2021learning} to project textual descriptions and rendered images derived from the 3D object into a modal-shared feature space. 
Subsequently, CLIP loss~\cite{jiang2023clip,wang2023seeing,ding2023clip4mc} is harnessed to align features from both modalities, optimizing the 3D representation to conform to the textual description.
Various scholars have made significant contributions to this field. 
For instance, Michel \emph{et al.}~\cite{michel2022text2mesh} are pioneers in proposing the use of CLIP loss to harmonize the text prompt with the rendered images of the 3D object, thereby enhancing text-to-3D generation.
Ma \emph{et al.}~\cite{ma2023x} introduce dynamic textual guidance during 3D object synthesis to improve convergence speed and generation performance.
However, these approaches have inherent limitations, as they tend to generate 3D representations with a scarcity of geometry and appearance detail.
To overcome this shortcoming, the diffusion-based text-to-3D approach~\cite{poole2022dreamfusion,lin2023magic3d,chen2023fantasia3d,tang2023make,jiang2023avatarcraft,huang2023dreamwaltz} leverages pretrained text-to-2D diffusion models~\cite{rombach2022high,saharia2022photorealistic} to guide the optimization of 3D representations.
Central to these models~\cite{cao2023texfusion,richardson2023texture,huang2024dreamwaltz,tang2023make,li2023diffusion} is the application of SDS loss~\cite{poole2022dreamfusion} to align the rendered images stemming from a variety of camera perspectives with the textual description.
Specifically, given the target text prompt, Lin \emph{et al.}~\cite{lin2023magic3d} leverage a coarse-to-fine pipeline to generate high-resolution 3D content.
Chen \emph{et al.}~\cite{chen2023fantasia3d} decouple geometry modeling and appearance modeling to generate realistic 3D assets.
For specific purposes, some researchers~\cite{seo2023let,wang2023prolificdreamer} integrate trainable LoRA~\cite{hu2021lora} branches into pretrained diffusion models. 
For instance, Seo \emph{et al.}~\cite{seo2023let} put forth 3DFuse, a model that harnesses the power of LoRA to comprehend object semantics. 
Wang \emph{et al.}~\cite{wang2023prolificdreamer} introduce ProlificDreamer, where the role of LoRA is to evaluate the score of the variational distribution for 3D parameters.
However, the LoRA parameter begins its journey from random initialization and maintains its independence from the camera and text.
To address these limitations,  we present two innovative modules: \moduleOneShort and \moduleTwoShort. These modules are designed to enhance the model's ability to consider important camera parameters and prioritize the generation of foreground objects throughout the text-to-3D creation process.

\begin{figure*}
  \centering
    % %\vspace{-1em}
  \includegraphics[width=1.0\columnwidth]{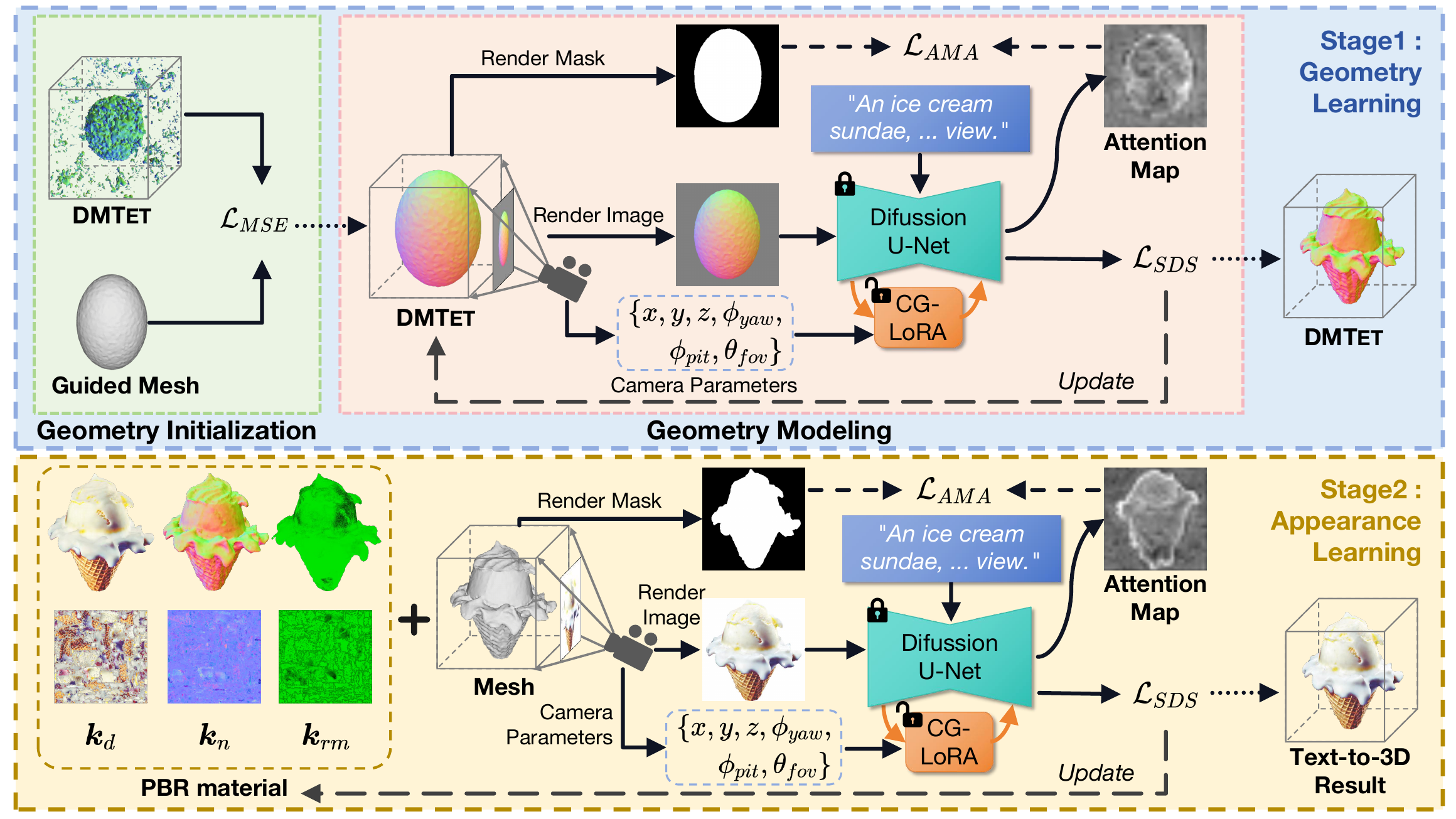}
    %\vspace{-0.5em}
  \caption{Overview of the proposed \modelname, which consists of geometry learning and appearance learning.}
  \label{fig:overview}
    % %\vspace{-1em}
\end{figure*}

\subsection{Low-Rank Adaptation (LoRA)}

Low-Rank Adaptation (LoRA)~\cite{hu2022lora} is a technique used to reduce memory requirements when fine-tuning a large model~\cite{wang2024visionllm,achiam2023gpt,li2024llava,chen2023internvl,liu2024visual,luo2024feast,li2023monkey}.
It involves injecting only a small set of 
trainable parameters into the pretrained model, while keeping the original parameters fixed. 
During the optimization process, gradients are passed through the fixed pretrained model weights to the LoRA adapter, which is then updated to optimize the loss function.
LoRA has been applied in various fields, including natural language processing~\cite{hu2021lora,brown2020language}, image synthesis~\cite{zhang2023adding} and 3D generation~\cite{seo2023let,wang2023prolificdreamer}. 
To achieve low-rank adaptation, a linear projection with a pretrained weight matrix $\mathbf{W}_0 \in \mathbb{R}^{d_{in} \times d_{out}}$ is augmented with an additional low-rank factorized projection.
This augmentation is represented as $\mathbf{W}_0 + \Delta\mathbf{W} = \mathbf{W}_0 + \mathbf{A}\mathbf{B}$, where $\mathbf{A} \in \mathbb{R}^{d_{in} \times r}$, $\mathbf{B} \in \mathbb{R}^{r \times d_{out}}$, and $r \ll \min(d_{in}, d_{out})$.
During training, $\mathbf{W}_0$ remains fixed, while $\mathbf{A}$ and $\mathbf{B}$ are trainable.
The modified forward pass, given the original forward pass $\mathbf{Y} = \mathbf{X} \mathbf{W}_0$, can be formulated as follows:
\begin{equation}
\mathbf{Y} = \mathbf{X} \mathbf{W}_0 + \mathbf{X} \mathbf{A}\mathbf{B}.
\label{eq:lora}
\end{equation}
In this paper, we introduce \moduleOneShort, which involves the dynamic generation of trainable parameters for $\mathbf{A}$ based on camera information. This technique allows for integrating perspective information, including camera parameters and direction-aware descriptions, into the pretrained text-to-2D diffusion model. As a result, our method significantly enhances text-to-3D generation capabilities.

\section{Approach}

\subsection{Architecture}

% 先介绍包含第一和第二阶段，第一阶段先初始化再进行集合建模；第二阶段进行外观建模

In this section, we present a comprehensive introduction to the proposed \modelname, which consists of two main stages: geometry learning and appearance learning.
For geometry learning, we employ \dmtet~\cite{shen2021deep} as the 3D representation. \dmtet is an MLP parameterized with $\Phi_{dmt}$ and is initialized with a 3D ellipsoid using the mean squared error (MSE) loss $\mathcal{L}_{MSE}$. Subsequently, we optimize \dmtet and \moduleOneShort using the SDS loss~\cite{poole2022dreamfusion} $\mathcal{L}_{SDS}$ and the proposed AMA loss $\mathcal{L}_{AMA}$ to ensure the alignment between the 3D representation and the input text prompt.
For appearance learning, we leverage bidirectional reflectance distribution function (BRDF) modeling~\cite{torrance1967theory} following the previous approach~\cite{chen2023fantasia3d}. Specifically, we utilize an MLP with trainable parameters $\Phi_{mat}$ to predict surface materials. Similar to the geometry learning stage, we optimize $\Phi_{mat}$ and \moduleOneShort using the SDS loss $\mathcal{L}_{SDS}$ and the AMA loss $\mathcal{L}_{AMA}$ to achieve alignment between the 3D representation and the text prompt.
\figref{fig:overview} provides a detailed depiction of our proposed \modelname.

\subsubsection{Geometry Learning}
% \noindent{\textbf{Geometry Learning.}}
% geometry initialization
For geometry learning, an MLP network $\Phi_{dmt}$ is utilized to parameterize \dmtet as a 3D representation. To enhance the stability of geometry modeling, we employ a 3D ellipsoid as the initial configuration for \dmtet $\Phi_{dmt}$.
For each vertex $v_i \in V_T$ belonging to the tetrahedral grid $T$, we train $\Phi_{dmt}$ to predict two important values: the SDF value $s(v_i)$ and the deformation offset $\delta(v_i)$.
To initialize $\Phi_{dmt}$ with the 3D ellipsoid, we sample a set of $N$ points $\{p_i \in \mathbb{R}^3\}|_{i=1}^N$ approximately distributed on the surface of an ellipsoid and compute the corresponding SDF values $\{{SDF}(p_i)\}|_{i=1}^N$.
Subsequently, we optimize $\Phi_{dmt}$ using MSE loss. This optimization process ensures that $\Phi_{dmt}$ effectively initializes \dmtet to resemble the 3D ellipsoid. The formulation of the MSE loss is given by:
\begin{equation}
\mathcal{L}_{MSE} = \frac{1}{N} \sum_{i=1}^N \left( s(p_i;\Phi_{dmt})-SDF(p_i) \right)^2.
\label{eq:mse}
\end{equation}
% geometry modeling
After initializing the geometry, our objective is to align the geometry of \dmtet with the input text prompt. Specifically, we generate the normal map $\boldsymbol{n}$ and the object mask $\boldsymbol{m}$ from the initialized \dmtet $\Phi_{dmt}$ by employing a differentiable rendering technique~\cite{torrance1967theory}, given a randomly sampled camera pose $\boldsymbol{c}$. Subsequently, we input the normal map $\boldsymbol{n}$ into the frozen stable diffusion (SD) with a trainable \moduleOneShort and update $\Phi_{dmt}$ using the SDS loss, which is defined as follows:
% {\small
\begin{equation}
\small
\nabla_{\Phi_{dmt}}\mathcal{L}_{\mathrm{SDS}} 
= \mathbb{E}_{t,\boldsymbol{\epsilon}}\left[w(t)\left(\hat{\boldsymbol{\epsilon}}_\Theta(\boldsymbol{n}_t;\boldsymbol{y},t)-\boldsymbol{\epsilon}\right)\frac{\partial \boldsymbol{n}}{\partial{\Phi_{dmt}}}\right],
\label{eq:sds}
\end{equation}
% }
where $\Theta$ represents the parameter of SD, $\hat{\boldsymbol{\epsilon}}_\Theta(\boldsymbol{n}_t;\boldsymbol{y},t)$ denotes the predicted noise of SD given the noise level $t$ and text embedding $\boldsymbol{y}$. 
Additionally, $\boldsymbol{n}_t=\alpha_t\boldsymbol{n}+\sigma_t\boldsymbol{\epsilon}$, where $\boldsymbol{\epsilon}\sim\mathcal{N}(\mathbf{0},\boldsymbol{I})$ represents noise sampled from a normal distribution.
The implementation of $w(t)$, $\alpha_t$, and $\sigma_t$ is based on the DreamFusion~\cite{poole2022dreamfusion}.

Furthermore, to focus SD on generating foreground objects, we introduce an additional AMA loss to align the object mask $\boldsymbol{m}$ with the attention map of SD, given by:
\begin{equation}
\mathcal{L}_{AMA} = \frac{1}{L}  \sum_{i=1}^L |\boldsymbol{a}_i - \eta(\boldsymbol{m})|,
\label{eq:ama}
\end{equation}
where $L$ denotes the number of attention layers, and $\boldsymbol{a}_i$ is the attention map of $i$-th attention layer. The function $\eta(\cdot)$ is employed to resize the rendered mask, ensuring its dimensions align with those of the attention maps.

\begin{figure}
  \centering
  \includegraphics[width=0.6\columnwidth]{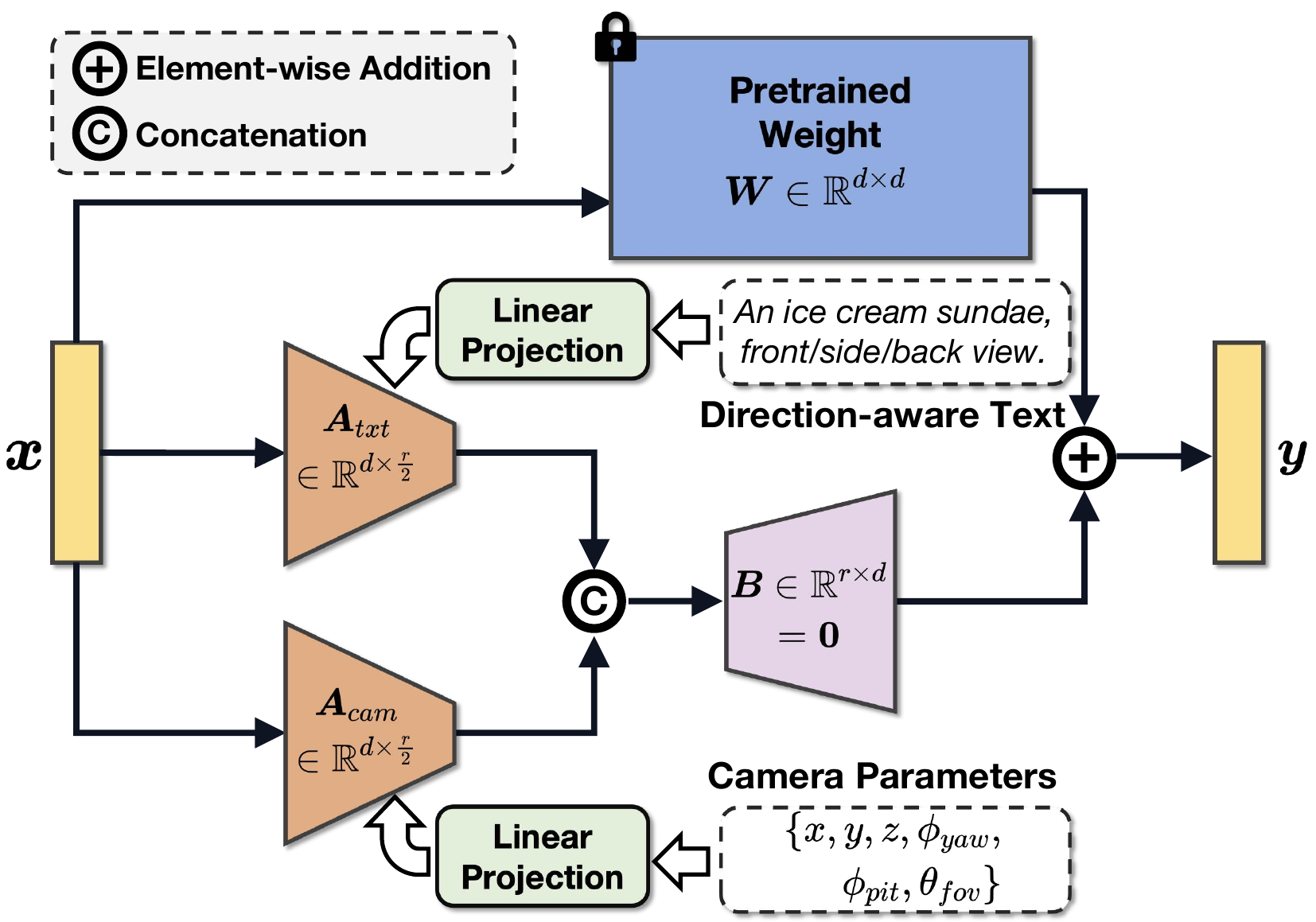}
  %\vspace{-1em}
  \caption{ Illustration of \moduleOneBig (\moduleOneShort).}
  \label{fig:cglora}
  %\vspace{-1em}
\end{figure}

\subsubsection{Appearance Learning}
% \noindent{\textbf{Appearance Learning.}}
After obtaining the geometry of the 3D object, our objective is to compute its appearance using the Physically-Based Rendering (PBR) material model~\cite{mcauley2012practical}. The material model comprises the diffuse term $\boldsymbol{k}_d \in \mathbb{R}^3$, the roughness and metallic term $\boldsymbol{k}_{rm} \in \mathbb{R}^2$, and the normal variation term $\boldsymbol{k}_{n} \in \mathbb{R}^3$. 
Firstly, the Diffuse Term $\boldsymbol{k}_d \in \mathbb{R}^3$ accurately represents how the material responds to diffuse lighting. This term is captured by a three-dimensional vector that denotes the material's diffuse color.
Secondly, the Roughness and Metallic Term $\boldsymbol{k}_{rm} \in \mathbb{R}^2$ effectively captures the material's roughness and metallic properties. It is represented by a two-dimensional vector that conveys the values of roughness and metallicness.
Roughness quantifies the surface smoothness, while metallicness indicates the presence of metallic properties in the material. These terms collectively contribute to the overall visual realism of the rendered scene.
Additionally, the Normal Variation Term $\boldsymbol{k}_n \in \mathbb{R}^3$ plays a crucial role in characterizing variations in surface normals. This term, typically used in conjunction with a normal map, enables the simulation of fine details and textures on the material's surface.
During the rendering process, these three PBR material terms interact to accurately simulate the optical properties exhibited by real-world materials. This integration facilitates rendering engines in achieving high-quality rendering results and generating realistic virtual scenes.

For any point $p \in \mathbb{R}^3$ on the surface of the geometry, we utilize an MLP parameterized by $\Phi_{mat}$ to obtain the three material terms, which can be expressed as follows:
\begin{equation}
(\boldsymbol{k}_d, \boldsymbol{k}_n, \boldsymbol{k}_{rm}) = \text{MLP}\left(\mathcal{P}(p); \Phi_{mat}\right),
\label{eq:pbr}
\end{equation}
where  $\mathcal{P}(\cdot)$ represents the positional encoding using a hash-grid technique~\cite{muller2022instant}. Subsequently, each pixel of the rendered image can be computed as follows:
\begin{equation}
V(p,\omega)=\int_{\Omega}L_i(p,\omega_i)f(p,\omega_i,\omega)(\omega_i\cdot n_p)\mathrm{d}\omega_i,
\label{eq:render}
\end{equation}
where $V(p,\omega)$ denotes the rendered pixel value from the direction $\omega$ for the surface point $p$.  $\Omega$ denotes a hemisphere defined by the set of incident directions $\omega_i$ satisfying the condition $\omega_i\cdot n_p \geq 0$, where $\omega_i$ denotes the incident direction, and $n_p$ represents the surface normal at point $p$. $L_i(\cdot)$ corresponds to the incident light from an off-the-shelf environment map, and $f(\cdot)$ is the Bidirectional Reflectance Distribution Function (BRDF) related to the material properties (\emph{i.e.,} $\boldsymbol{k}_d, \boldsymbol{k}_n, \boldsymbol{k}_{rm}$). 
By aggregating all rendered pixel colors, we obtain a rendered image $\boldsymbol{x} = \{ V(p,\omega) \}$. Similar to the geometry modeling stage, we feed the rendered image $\boldsymbol{x}$ into SD.
The optimization objective remains the same as \equref{eq:sds} and \equref{eq:ama}, where the rendered normal map $\boldsymbol{n}$ and the parameters of \dmtet $\Phi_{dmt}$ are replaced with the rendered image $\boldsymbol{x}$ and the parameters of the material encoder $\Phi_{mat}$, respectively.

\subsection{\moduleOneBig}

The domain gap between text-to-2D and text-to-3D generation presents a significant challenge, as discussed in \secref{sec:intro}, which leads to unsatisfied generated results.
To address these issues, we propose \moduleOneBig (\moduleOneShort) as a solution to bridge the domain gap.
As depicted in \figref{fig:cglora}, we leverage camera parameters and direction-aware text to guide the generation of parameters in \moduleOneShort, enabling \modelname to effectively incorporate camera perspective and direction information.

Specifically, given a text prompt $T$ and camera parameters $C=\{x, y, z, \phi_{yaw}, \phi_{pit}, \theta_{fov}\}$~\footnote{ The variables $x, y, z, \phi_{yaw}, \phi_{pit}, \theta_{fov}$ represent the x, y, z coordinates, yaw angle, pitch angle of the camera, and field of view, respectively. The roll angle $\phi_{roll}$ is intentionally set to 0 to ensure the stability of the object in the rendered image.}, we initially project these inputs into a feature space using the pretrained textual CLIP encoder $\mathcal{E}_{txt}(\cdot)$ and a trainable MLP $\mathcal{E}_{pos}(\cdot)$:
\begin{align}
\boldsymbol{t} &= \mathcal{E}_{txt}(T), \\
\boldsymbol{c} &= \mathcal{E}_{pos}(C),
\label{eq:project}
\end{align}
where $\boldsymbol{t} \in \mathbb{R}^{d_{txt}}$ and $\boldsymbol{c} \in \mathbb{R}^{d_{cam}}$ are textual features and camera features. Subsequently, we employ two low-rank matrices to project $\boldsymbol{t}$ and $\boldsymbol{c}$ into trainable dimensionality-reduction matrices within \moduleOneShort:
\begin{align}
\boldsymbol{A}_{txt} &= \text{Reshape}(\boldsymbol{t} \boldsymbol{W}_{txt}), \\
\boldsymbol{A}_{cam} &= \text{Reshape}(\boldsymbol{c} \boldsymbol{W}_{cam}),
\label{eq:cglora}
\end{align}
where $\boldsymbol{A}_{txt} \in \mathbb{R}^{d \times \frac{r}{2}}$ and $\boldsymbol{A}_{cam} \in \mathbb{R}^{d \times \frac{r}{2}}$ are two dimensionality-reduction matrices of \moduleOneShort.
The function $\text{Reshape}(\cdot)$ is used to transform the shape of a tensor from $\mathbb{R}^{d * \frac{r}{2}}$ to $\mathbb{R}^{d \times \frac{r}{2}}$. 
\footnote{$\mathbb{R}^{d * \frac{r}{2}}$ denotes a one-dimensional vector. $\mathbb{R}^{d \times \frac{r}{2}}$ represents a two-dimensional matrix.}
$\boldsymbol{W}_{txt} \in \mathbb{R}^{d_{txt} \times \left( d*\frac{r}{2} \right) }$ and $\boldsymbol{W}_{cam} \in \mathbb{R}^{d_{cam} \times \left( d * \frac{r}{2} \right)}$ are two low-rank matrices. Thus, we decompose them into the product of two matrices to reduce the trainable parameters in our implementation, \emph{i.e.,} $\boldsymbol{W}_{txt}=\boldsymbol{U}_{txt} \boldsymbol{V}_{txt}$ and $\boldsymbol{W}_{cam}=\boldsymbol{U}_{cam} \boldsymbol{V}_{cam}$, where $\boldsymbol{U}_{txt} \in \mathbb{R}^{d_{txt} \times r^\prime}$, 
$\boldsymbol{V}_{txt} \in  \mathbb{R}^{r^\prime  \times (d*\frac{r}{2})}$,
$\boldsymbol{U}_{cam} \in \mathbb{R}^{d_{cam} \times r^\prime}$,
$\boldsymbol{V}_{cam} \in  \mathbb{R}^{r^\prime 
\times  (d*\frac{r}{2})}$, 
$r^\prime$ is a small number (\emph{i.e.,} 4). In accordance with LoRA~\cite{hu2021lora}, we initialize the dimensionality-expansion matrix $\boldsymbol{B} \in \mathbb{R}^{r \times d}$ with zero values to ensure that the model begins training from the pretrained parameters of SD. Thus, the feed-forward process of \moduleOneShort is formulated as follows:
\begin{align}
\boldsymbol{y} = \boldsymbol{x}\boldsymbol{W} + [\boldsymbol{x}\boldsymbol{A}_{txt}; \boldsymbol{x}\boldsymbol{A}_{cam}]\boldsymbol{B},
\label{eq:feedfward}
\end{align}
where $\boldsymbol{W} \in \mathbb{R}^{d \times d}$ represents the frozen parameters of the pretrained SD model, and $[\cdot;\cdot]$ is the concatenation operation alone the channel dimension. In our implementation, we integrate \moduleOneShort into the linear embedding layers of the attention modules in SD to effectively capture direction and camera information.

\subsection{\moduleTwoBig}
Although SD is pretrained to generate 2D images that encompass both foreground and background elements, the task of text-to-3D generation demands a stronger focus on generating foreground objects. 
To address this specific requirement, we introduce \moduleTwoSL, which aims to align the attention map of SD with the rendered mask image of the 3D object.
Specifically, for each attention layer in the pretrained SD, we compute the attention map between the query image feature $\boldsymbol{Q} \in \mathbb{R}^{H \times h \times w \times \frac{d}{H}}$ and the key \texttt{CLS} token feature $\boldsymbol{K} \in \mathbb{R}^{H \times \frac{d}{H}}$. The calculation is formulated as follows:
\begin{align}
\bar{\boldsymbol{a}}=\mathrm{Softmax}(\frac{\boldsymbol{Q}\boldsymbol{K}^\top}{\sqrt{d}}),
\label{eq:att}
\end{align}
where $H$ denotes the number of attention heads in SD. $h$, $w$, and $d$ represent the height, width, and channel dimensions of the image features, respectively. $\bar{\boldsymbol{a}} \in \mathbb{R}^{H \times h \times w}$ represents the attention map.
Subsequently, we proceed to compute the overall attention map $\hat{\boldsymbol{a}} \in \mathbb{R}^{h \times w}$ by averaging the attention values of $\bar{\boldsymbol{a}}$ across all attention heads.
Since the attention map values are normalized using the softmax function, the activation values in the attention map may become very small when the image feature resolution is high.
However, considering that each element in the rendered mask has a binary value of either 0 or 1, directly aligning the attention map with the rendered mask is not optimal.
To address this, we propose a normalization technique that maps the values in the attention map from 0 to 1. This normalization process is formulated as follows:
\begin{align}
\boldsymbol{a}=\frac{\hat{\boldsymbol{a}}-\min(\hat{\boldsymbol{a}})}{\max(\hat{\boldsymbol{a}})-\min(\hat{\boldsymbol{a}})+\nu},
\label{eq:norm}
\end{align}
where $\nu$ represents a small constant value (\emph{e.g.,} $1e\textrm{-}6$) that prevents division by zero in the denominator. Finally, we align the attention maps of all attention layers with the rendered mask of the 3D object using the  \moduleTwoShort. The formulation of this alignment is presented in \equref{eq:ama}.

\section{Experiments}

\subsection{Implementation Details.}
We conduct the experiments using four Nvidia RTX 3090 GPUs and the PyTorch library~\cite{paszke2019pytorch}. To calculate the SDS loss, we utilize the Stable Diffusion implemented by HuggingFace Diffusers~\cite{von-platen-etal-2022-diffusers}. For the \dmtet $\Phi_{dmt}$ and material encoder $\Phi_{mat}$, we implement them as a two-layer MLP and a single-layer MLP, respectively, with a hidden dimension of 32. The values of $d_{cam}$, $d_{txt}$, $r$, $r^\prime$, the batch size, the SDS loss weight, the AMA loss weight, and the aspect ratio of the perspective projection plane are set to 1024, 1024, 4, 4, 4, 1, 0.1, and 1 respectively.
\moduleTwoShort is used after half of the training iterations, where \moduleOneShort has a certain degree of perspective alignment ability.
We optimize \modelname for 2000 iterations for geometry learning and 1000 iterations for appearance learning. 
For each iteration, $\phi_{pit}$, $\phi_{yaw}$, and $\theta_{fov}$ are randomly sampled from $(-15^\circ, 45^\circ)$, $(-180^\circ, 180^\circ)$, and $(25^\circ, 45^\circ)$, respectively.
%The resolution of the rendered image during training is $512 \times 512$.

\begin{figure*}
  \centering
  \includegraphics[width=1.0\columnwidth]{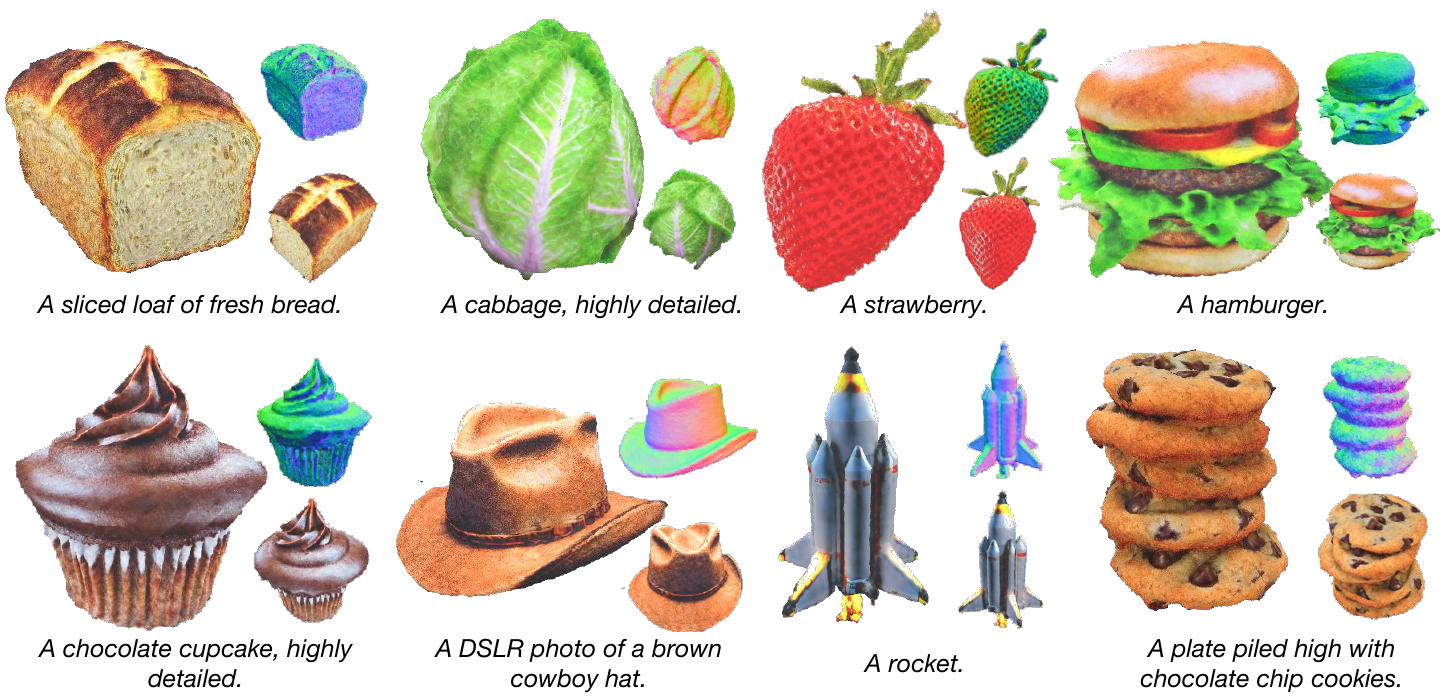}
    %\vspace{-1.0em}
  \caption{Text-to-3D generation results from an ellipsoid.}
  \label{fig:textto3D1}
    %\vspace{-0.5em}
\end{figure*}

\begin{figure*}
  \centering
  \includegraphics[width=1.0\columnwidth]{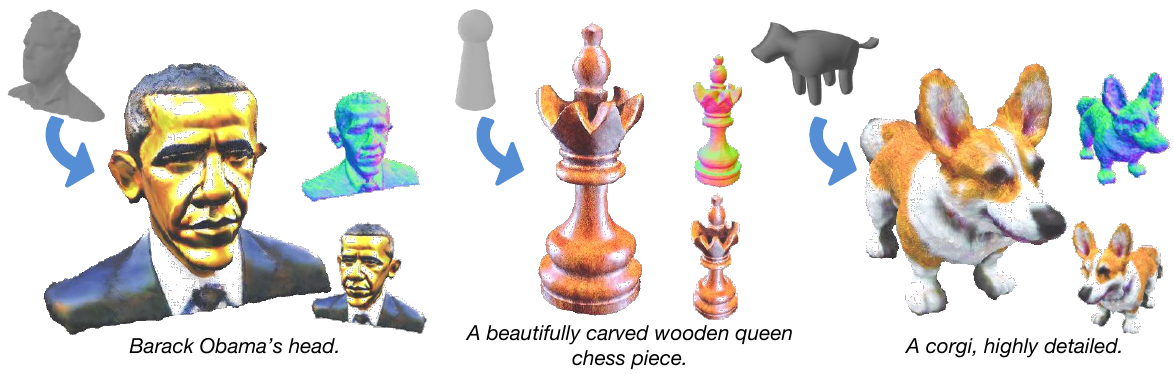}
    %\vspace{-1.0em}
  \caption{Text-to-3D generation results from coarse-grained guided meshes.}
  \label{fig:textto3D2}
    %\vspace{-1em}
\end{figure*}

\subsection{Results of \modelname}

\subsubsection{Text-to-3D generation from an ellipsoid}
% \noindent\textbf{Text-to-3D generation from an ellipsoid.}
We present representative results of \modelname for text-to-3D generation, utilizing an ellipsoid as the initial geometry, as shown in \figref{fig:textto3D1}.
The results demonstrate the ability of \modelname to generate high-quality and photo-realistic outputs that accurately correspond to the input text prompts.

\subsubsection{Text-to-3D generation from coarse-grained meshes}
% \noindent\textbf{Text-to-3D generation from coarse-grained meshes.}
% 
While there is a wide availability of coarse-grained meshes for download from the internet, directly utilizing these meshes for 3D content creation often results in poor performance due to the lack of geometric details.
However, when compared to a 3D ellipsoid, these meshes may provide better 3D shape prior information for \modelname.
Hence, instead of using ellipsoids, we can initialize \dmtet with coarse-grained guided meshes as well.
As shown in \figref{fig:textto3D2}, \modelname can generate 3D assets with precise geometric details based on the given text, even when the provided coarse-grained mesh lacks details. 
For instance, in the last column of \figref{fig:textto3D2}, \modelname accurately transforms the geometry from a cow to a corgi based on the text prompt ``A corgi, highly detailed."
Therefore, \modelname is also an exceptionally powerful tool for editing coarse-grained mesh geometry using textual inputs.

\begin{table*}[]
\caption{Quantitative comparison of SOTA Methods: The top-performing and second-best results are highlighted in \textbf{bolded} and \underline{underlined}, respectively.}
% %\vspace{-1em}
\setlength{\tabcolsep}{3.0mm}{
\resizebox{\textwidth}{!}{
\begin{tabular}{l|cc|ccc|ccc}
\hline
\multirow{2}{*}{\textbf{Method}} & \multicolumn{2}{c|}{\textbf{User Study}} & \multicolumn{3}{c|}{\textbf{CLIP Score}}      & \multicolumn{3}{c}{\textbf{OpenCLIP Score}}   \\ \cline{2-9} 
                                 & Geo. Qua.           & App. Qua.          & ViT-B/32      & ViT-B/16      & ViT-L/14      & ViT-B/32      & ViT-B/16      & ViT-L/14      \\ \hline
Dreamfusion~\cite{poole2022dreamfusion}                     & 1.2                 & 0.6                & 30.5          & 31.1          & 26.3          & 31.9          & 27.9          & 30.0          \\
Magic3d~\cite{lin2023magic3d}                          & 2.1                 & 1.1                & 28.4          & 28.6          & 24.6          & 29.1          & 25.9          & 28.3          \\
Fantasia3d~\cite{chen2023fantasia3d}                       & 5.4                 & 6.3                & 30.4          & 30.1          & 24.8          & 30.3          & 27.2          & 29.7          \\
ProlificDreamer~\cite{wang2023prolificdreamer}                  & {\ul 12.9}          & {\ul 16.2}         & {\ul 30.8}    & {\ul 31.5}    & {\ul 26.4}    & {\ul 32.7}    & {\ul 28.8}    & {\ul 31.1}    \\
X-Dreamer                         & \textbf{78.4}       & \textbf{75.8}      & \textbf{32.0} & \textbf{32.2} & \textbf{27.2} & \textbf{34.1} & \textbf{29.6} & \textbf{32.2} \\
\hline
\end{tabular}
}
}
\label{tab:score}
\end{table*}

\begin{figure*}
  \centering
  % %\vspace{-1.0em} 
  \includegraphics[width=1.0\columnwidth]{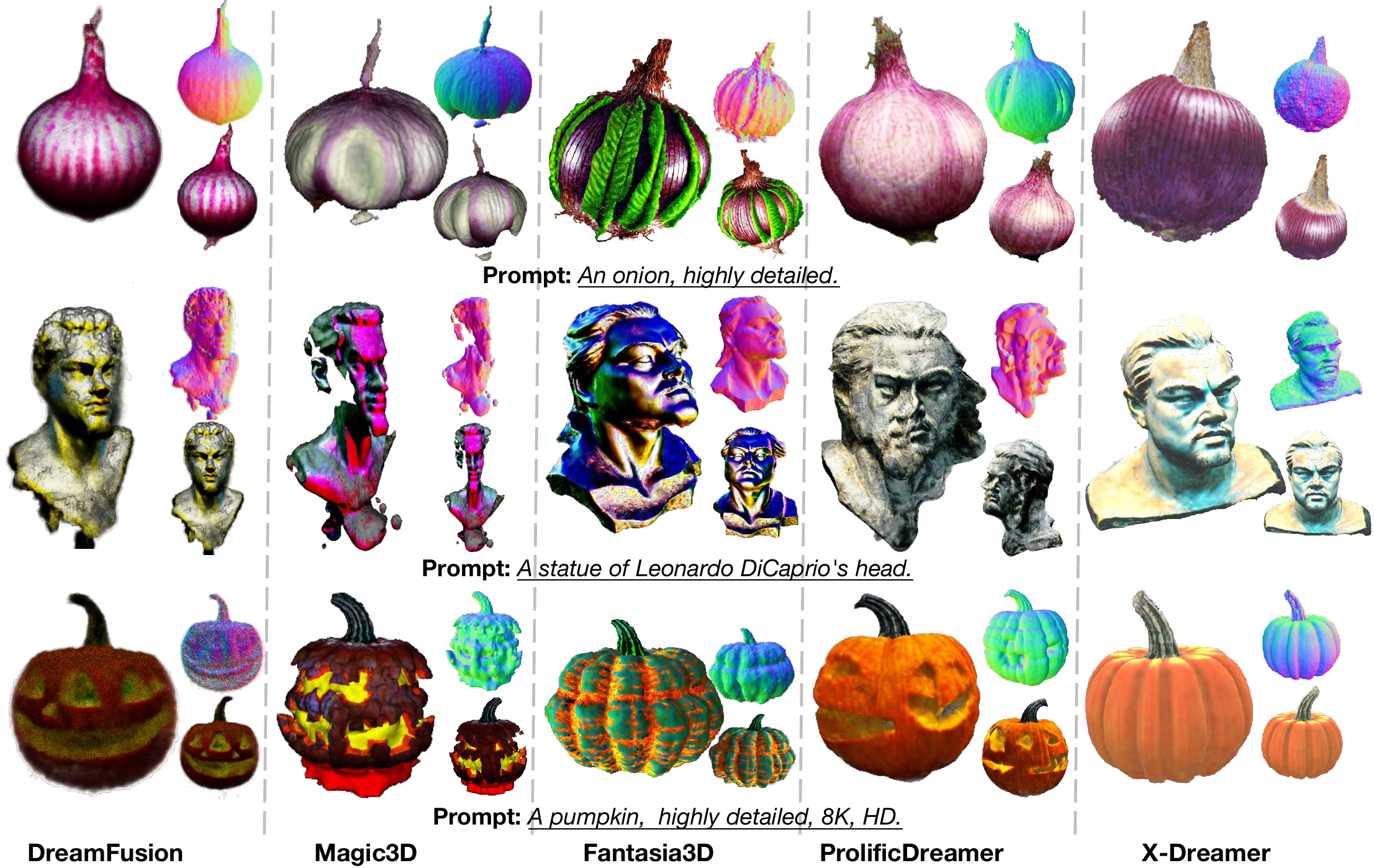}
    % %\vspace{-1.0em}
  \caption{Comparison with SOTA methods. Our method yields results that exhibit enhanced fidelity and more details.}
  \label{fig:comparison}
    %\vspace{-1em}
\end{figure*}

\begin{figure*}
  \centering
  \includegraphics[width=1.0\columnwidth]{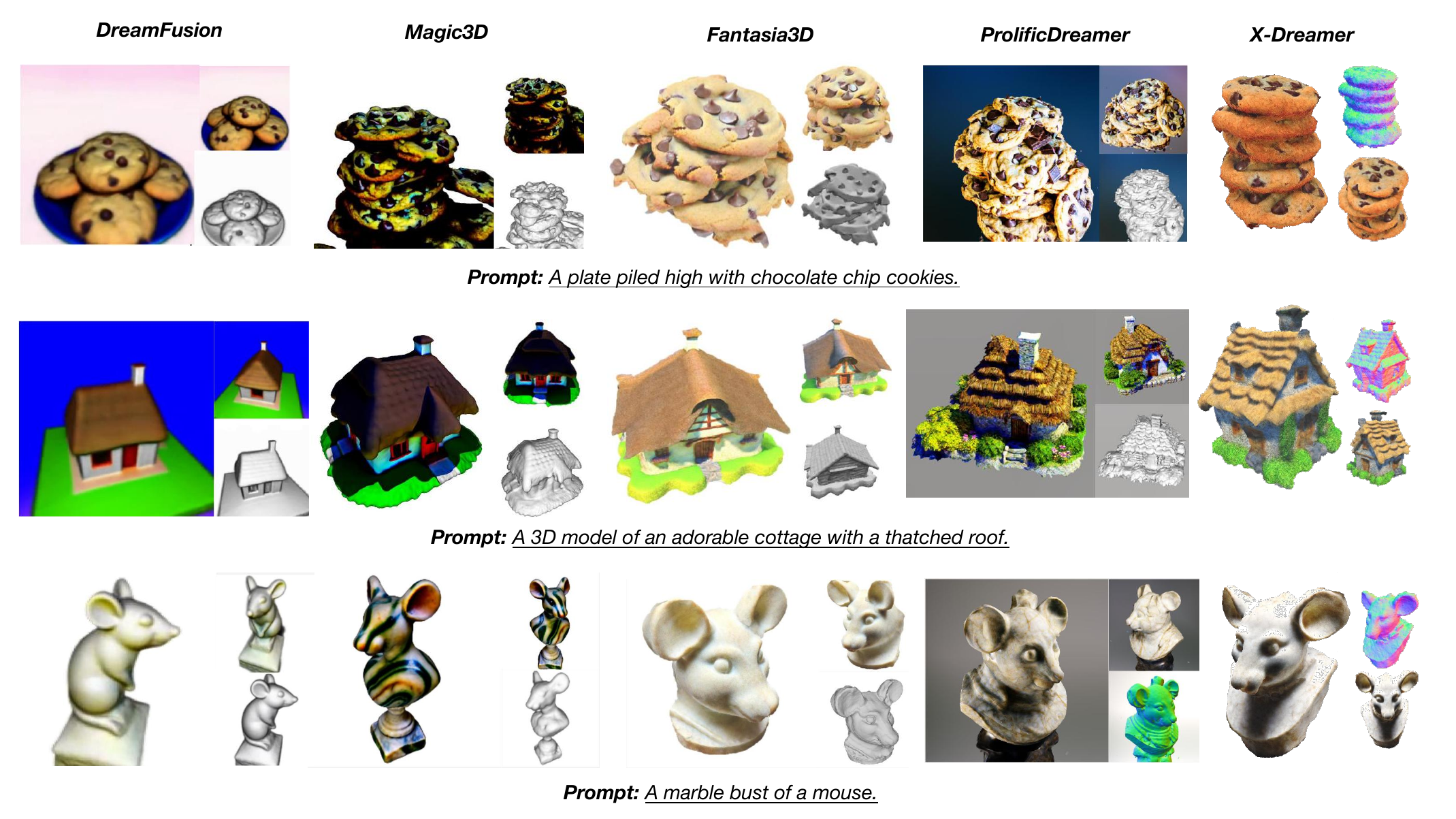}
  \caption{More comparison with State-of-the-Art (SOTA) methods.}
  \label{fig:more_comparison_paper}
\end{figure*}

\subsubsection{Quantitative Comparison}
% \noindent\textbf{Quantitative Comparison.}
We conducted a meticulously organized user study involving 50 individuals to evaluate the performance of four state-of-the-art (SOTA) methods by assessing 50 generated results, \emph{i.e.,} DreamFusion~\cite{poole2022dreamfusion}, Magic3D~\cite{lin2023magic3d}, Fantasia3D~\cite{chen2023fantasia3d}, and ProlificDreamer~\cite{wang2023prolificdreamer}. Participants were asked to compare and select the best result based on two criteria: Geometry Quality (Geo. Qua.) and Appearance Quality (App. Qua.). The results, as presented in Tab.~\ref{tab:score}, indicate that our approach garnered a preference from 75.0\% of the participants. This overwhelming preference underscores our method's exceptional quality and superiority, establishing it as a pioneering solution in the field.
The primary objective of these user studies was to evaluate the geometric and visual quality of the generated results, without specifically assessing their alignment with the text. To provide a more comprehensive evaluation, we further calculated the CLIP score and OpenCLIP score for both the generated results and the corresponding text prompts. As depicted in Tab.~\ref{tab:score}, our method outperformed other approaches in terms of both the CLIP score and OpenCLIP score. These findings demonstrate that the results generated by our method exhibit superior alignment with the text, further strengthening the efficacy of our approach.

\subsubsection{Qualitative Comparison}
% \noindent\textbf{Qualitative Comparison.}
To assess the effectiveness of \modelname, we compare it with four SOTA methods. Since the codes of some methods~\cite{poole2022dreamfusion,lin2023magic3d,wang2023prolificdreamer} have not been publicly available, we present the results obtained by implementing these methods in threestudio~\cite{threestudio2023}. The results are depicted in \figref{fig:comparison}.
In \figref{fig:more_comparison_paper}, we present a comprehensive comparison of our methods with four baselines, using the images provided in their original papers.
When compared to the SDS-based methods~\cite{poole2022dreamfusion,lin2023magic3d,chen2023fantasia3d}, \modelname outperforms them in generating superior-quality and realistic 3D assets.
In addition, when compared to the VSD-based method~\cite{wang2023prolificdreamer}, \modelname produces 3D content with comparable or even better visual effects, while requiring significantly less optimization time.
Specifically, the geometry and appearance learning process of \modelname requires only approximately 27 minutes, whereas ProlificDreamer exceeds 8 hours.

\begin{figure}
  \centering
  \includegraphics[width=1.0\columnwidth]{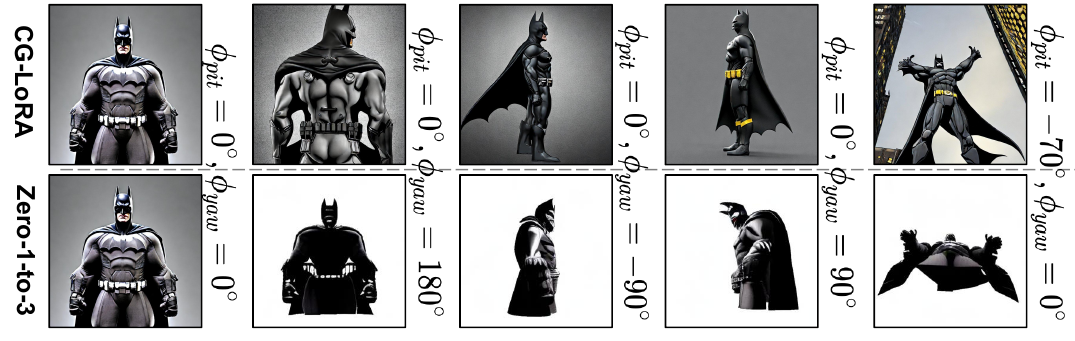}
  %\vspace{-2em}
  \caption{ Camera Control Comparison with Zero-1-to-3~\cite{liu2023zero}.}
  \label{fig:angle}
  %\vspace{-1em}
\end{figure}

\subsubsection{Camera Control Comparison}
% \noindent\textbf{Camera Control Comparison.}
% 
During training, \moduleOneShort aims to control the camera parameters. To investigate the effectiveness of this control, we visualize the image generated by Stable Diffusion (SD) with a pretrained \moduleOneShort. As shown in Fig.~\ref{fig:angle}, given a specific camera parameter, SD with \moduleOneShort can generate images from that perspective. To make a comparative analysis, we also compare our results with Zero-1-to-3~\cite{liu2023zero}, a method capable of generating images with desired viewing angles based on relative camera angles and reference images. Using the front view produced by our approach as a reference image, we leverage Zero-1-to-3 to generate images with varying camera parameters. Our observations indicate that our method can generate images that exhibit a general alignment with the angles produced by Zero-1-to-3.

\subsection{Ablation Study}

\begin{figure*}
    % %\vspace{-1.0em}
  \centering
  \includegraphics[width=1.0\columnwidth]{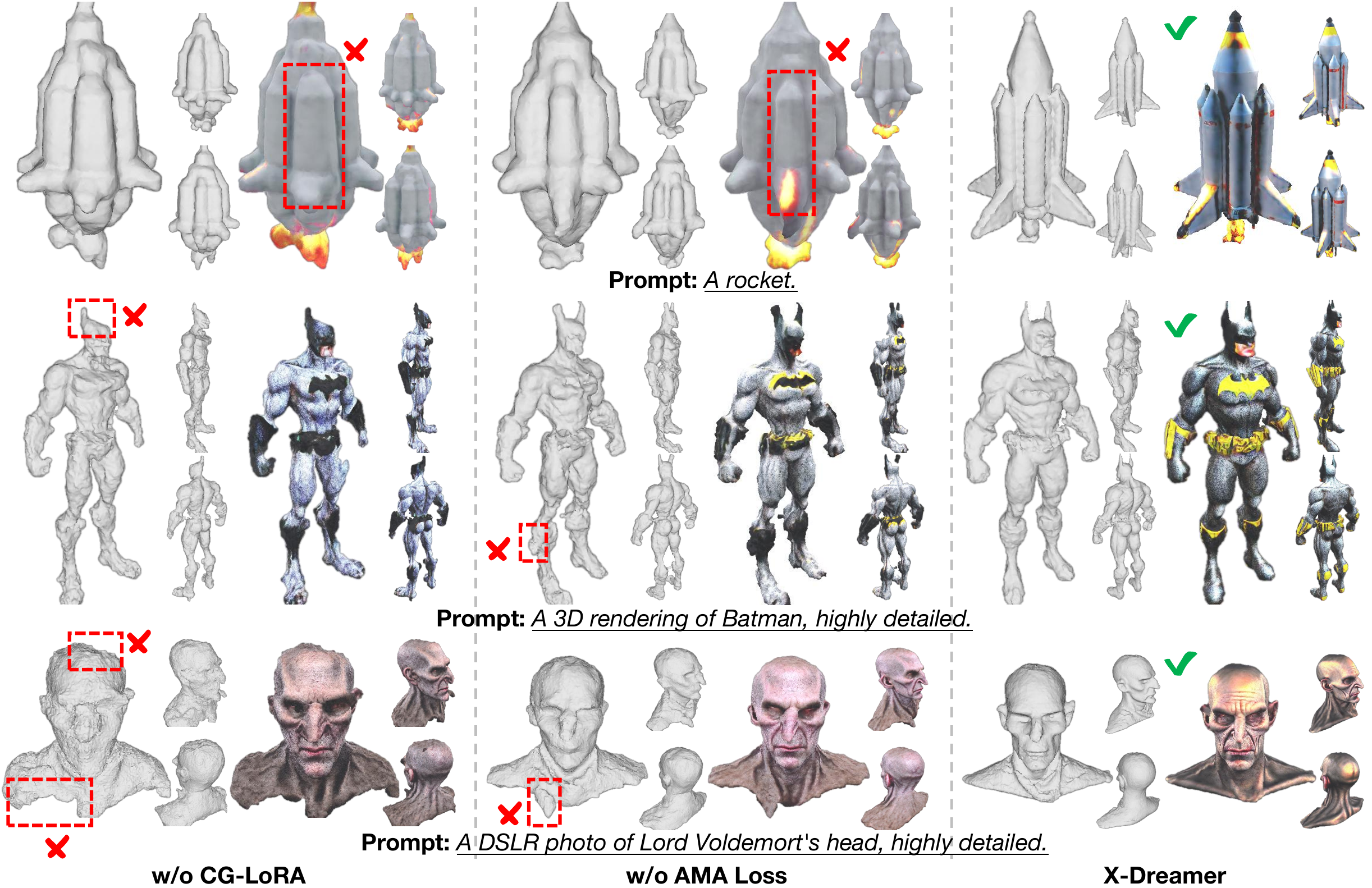}
    % %\vspace{-0.9em}
  \caption{Ablation studies of the proposed \modelname.}
  \label{fig:ablation}
    %\vspace{-0.5em}
\end{figure*}

\begin{figure*}
  \centering
  \includegraphics[width=1.0\columnwidth]{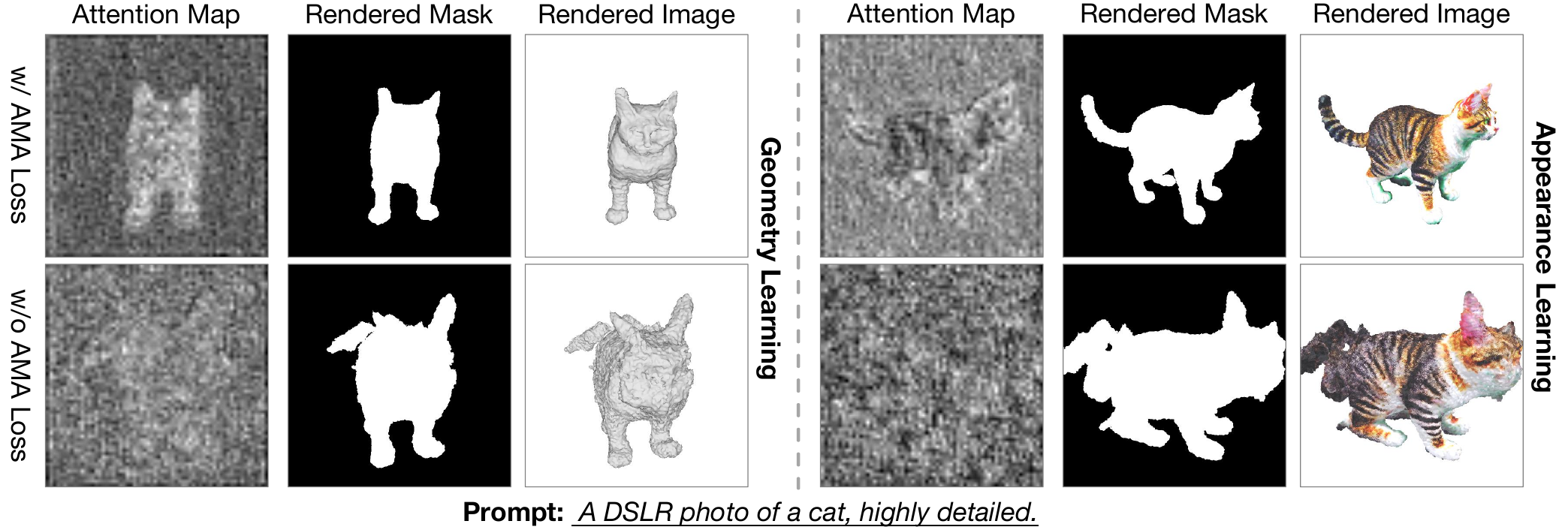}
    % %\vspace{-1.0em}
  \caption{Visualization of Attention Map, Rendered Mask, and Rendered Image with and without AMA Loss. For clarity, we only visualize the attention map of the first attention layer in SD.}
  \label{fig:amaloss}
    % %\vspace{-1.0em}
\end{figure*}

\subsubsection{Ablation on the proposed modules}
% \noindent\textbf{Ablation on the proposed modules.}
% 放可视化的结果对比
To gain insights into the abilities of \moduleOneShort and \moduleTwoShort, we perform ablation studies wherein each module is incorporated individually to assess its impact.
As depicted in \figref{fig:ablation}, the ablation results demonstrate a notable decline in the geometry and appearance quality of the generated 3D objects when \moduleOneShort is excluded from \modelname.
For instance, as shown in the second row of \figref{fig:ablation}, the generated Batman lacks an ear on the top of its head in the absence of \moduleOneShort.
This observation highlights the crucial role of \moduleOneShort in injecting camera-relevant information into the model, thereby enhancing the 3D consistency.
Furthermore, the omission of \moduleTwoShort from \modelname also has a deleterious effect on the geometry and appearance fidelity of the generated 3D assets.
Specifically, as illustrated in the first row of \figref{fig:ablation}, \modelname successfully generates a photorealistic texture for the rocket, whereas the texture quality noticeably deteriorates in the absence of \moduleTwoShort.
This disparity can be attributed to \moduleTwoShort, which directs the focus of the model towards foreground object generation, ensuring the realistic representation of both geometry and appearance of foreground objects.
These ablation studies provide valuable insights into the individual contributions of \moduleOneShort and \moduleTwoShort in enhancing the geometry, appearance, and overall quality of the generated 3D objects.

\subsubsection{Attention map comparisons w/ and w/o \moduleTwoShort}
% \noindent\textbf{Attention map comparisons w/ and w/o \moduleTwoShort.}
% 放Attention Map的对比
\moduleTwoShort is introduced with the aim of guiding attention during the denoising process towards the foreground object.
This objective is achieved by aligning the attention map of SD with the rendered mask of the 3D object.
To evaluate the effectiveness of \moduleTwoShort in accomplishing this goal, we visualize the attention maps of SD with and without \moduleTwoShort at both the geometry learning and appearance learning stages.
As depicted in \figref{fig:amaloss}, it can be observed that incorporating \moduleTwoShort not only results in improved geometry and appearance of the generated 3D asset, but also concentrates the attention of SD specifically on the foreground object area.
The visualizations confirm the efficacy of \moduleTwoShort in directing the attention of SD, resulting in improved quality and foreground object focus during geometry and appearance learning stages.

\subsubsection{Ablation on the parameter generation of \moduleOneShort}

\begin{figure*}
  \centering
  \includegraphics[width=1.0\columnwidth]{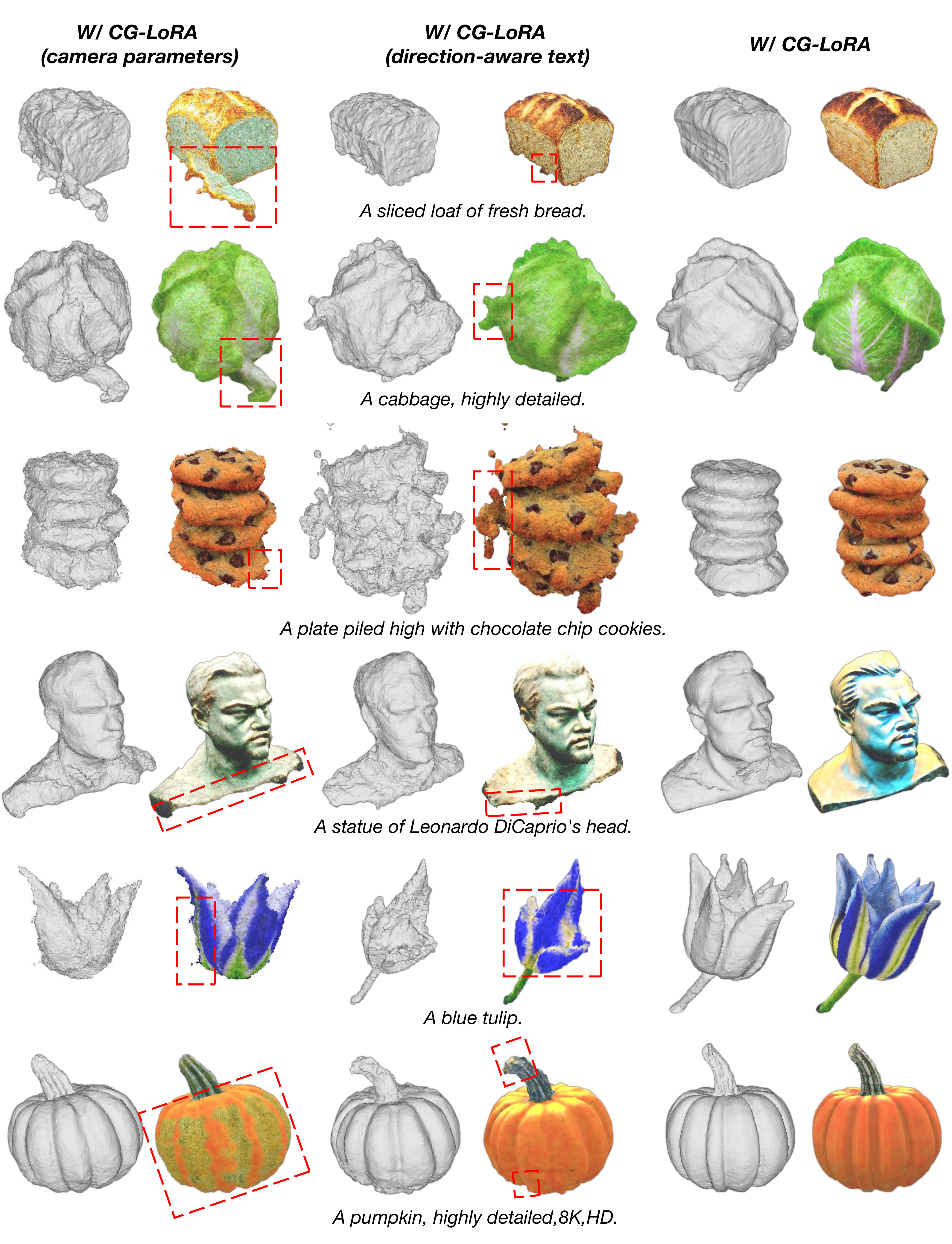}
  \caption{Ablation Study of \moduleOneShort.}
  \label{fig:abcglora}
\end{figure*}

In our implementation, the parameters of \moduleOneShort are dynamically generated based on two key factors related to camera information: camera parameters and direction-aware text.
To thoroughly investigate the impact of these camera information terms on \modelname, we conducted ablation experiments, dynamically generating CG-LoRA parameters based solely on one of these terms.
The results, as illustrated in \figref{fig:abcglora}, clearly demonstrate that when using only one type of camera information term, the geometry and appearance quality of the generated results are significantly diminished compared to those produced by \modelname.
For example, as shown in the third row of \figref{fig:abcglora}, when utilizing the \moduleOneShort solely related to direction-aware text, the geometry quality of the generated cookies appears poor. Similarly, as depicted in the last row of \figref{fig:abcglora}, employing the \moduleOneShort solely based on camera parameters results in diminished geometry and appearance quality for the pumpkin.
These findings highlight the importance of both camera parameters and direction-aware text in generating high-quality 3D assets. The successful generation of parameters for \moduleOneShort relies on the synergy between these two camera information terms, emphasizing their crucial role in achieving superior geometry and appearance in the generated assets.

\subsection{Attention Map of Different Layers}

\begin{figure*}
  \centering
  \includegraphics[width=1.0\columnwidth]{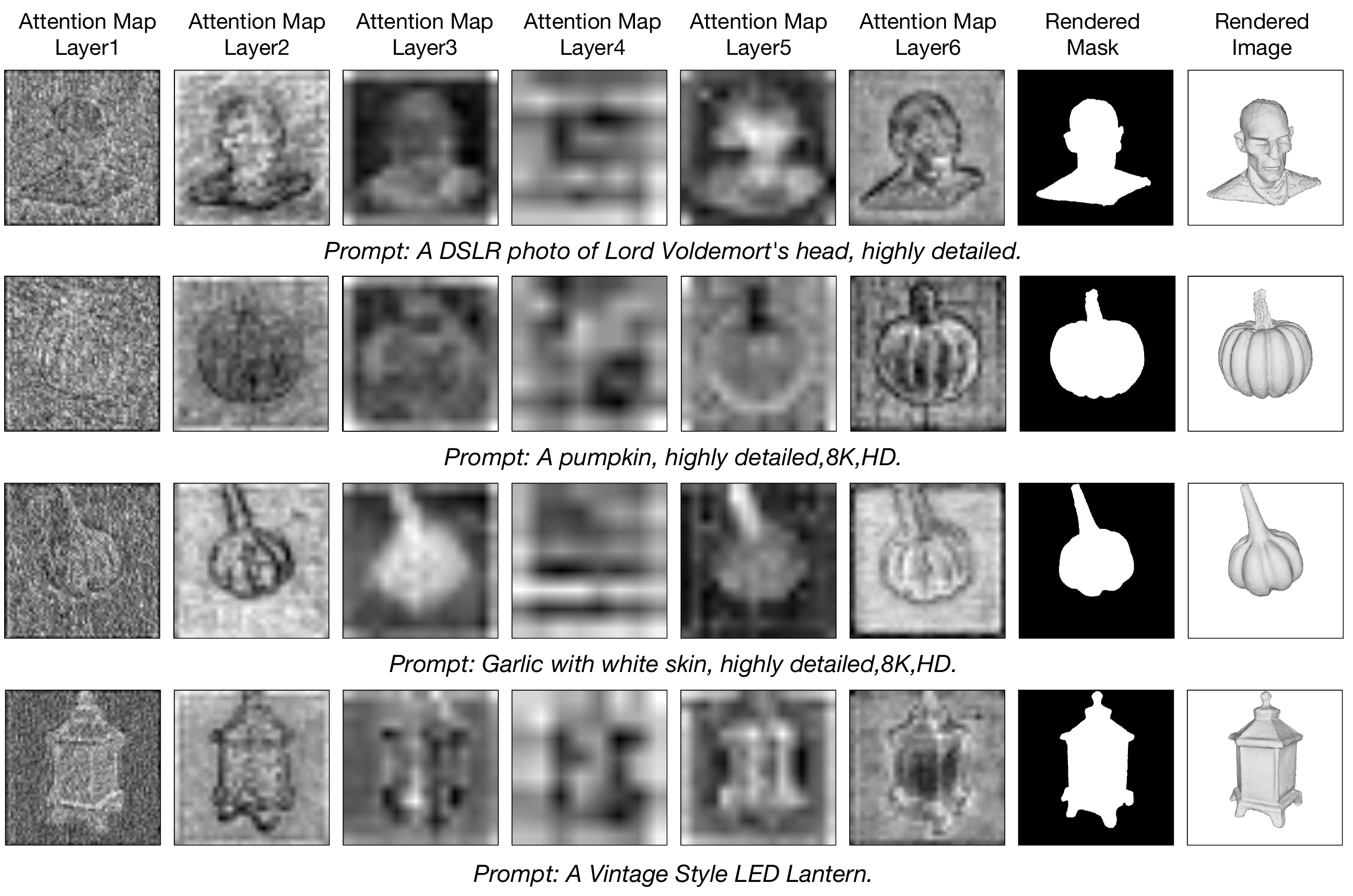}
  \caption{Attention maps of different layers of SD for the geometry learning stage.}
  \label{fig:att_layers_geo}
\end{figure*}

\begin{figure*}
  \centering
  \includegraphics[width=1.0\columnwidth]{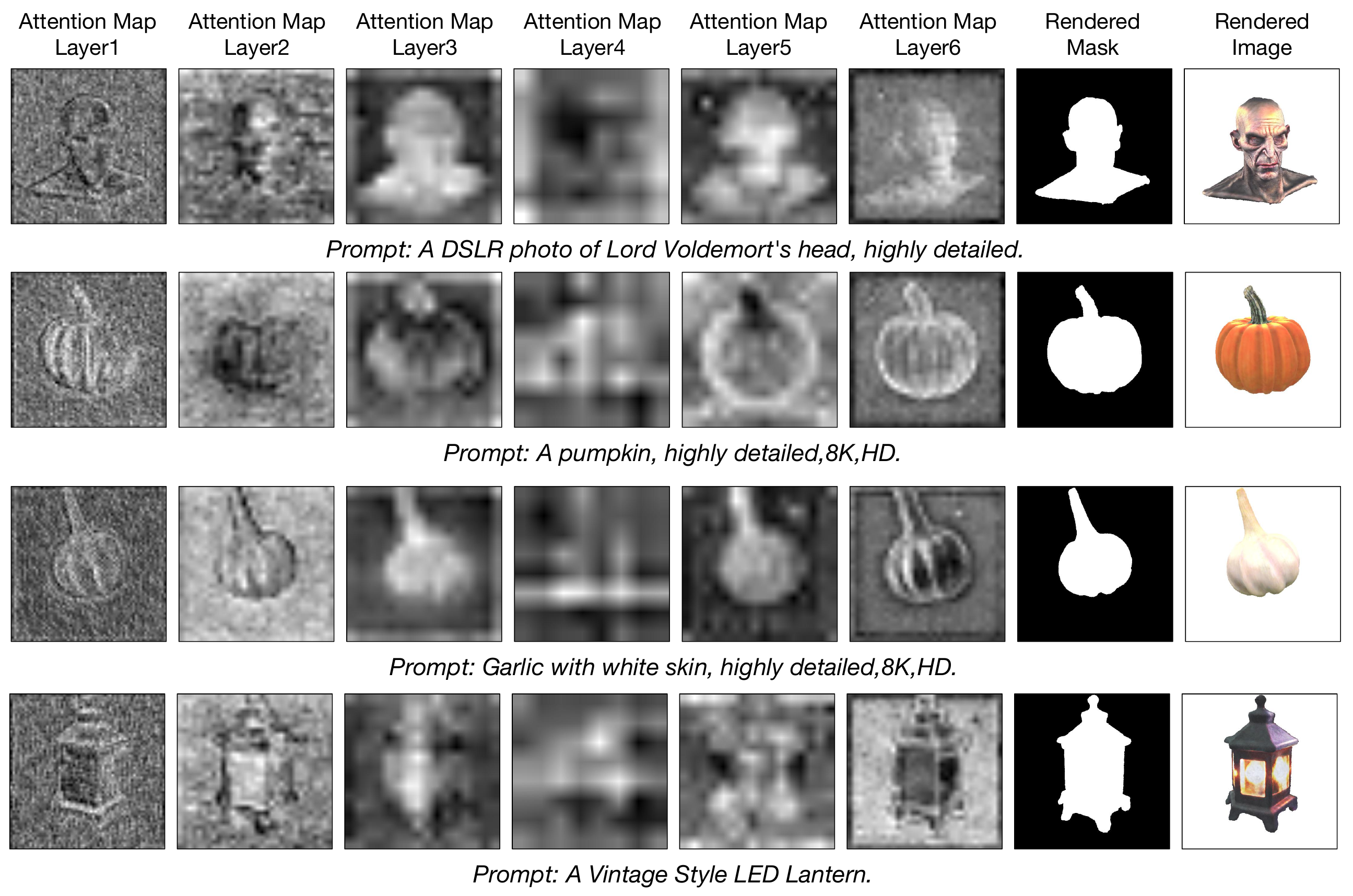}
  \caption{Attention maps of different layers of SD for the appearance learning stage.}
  \label{fig:att_layers_app}
\end{figure*}

\begin{figure*}
  \centering
  \includegraphics[width=1.0\columnwidth]{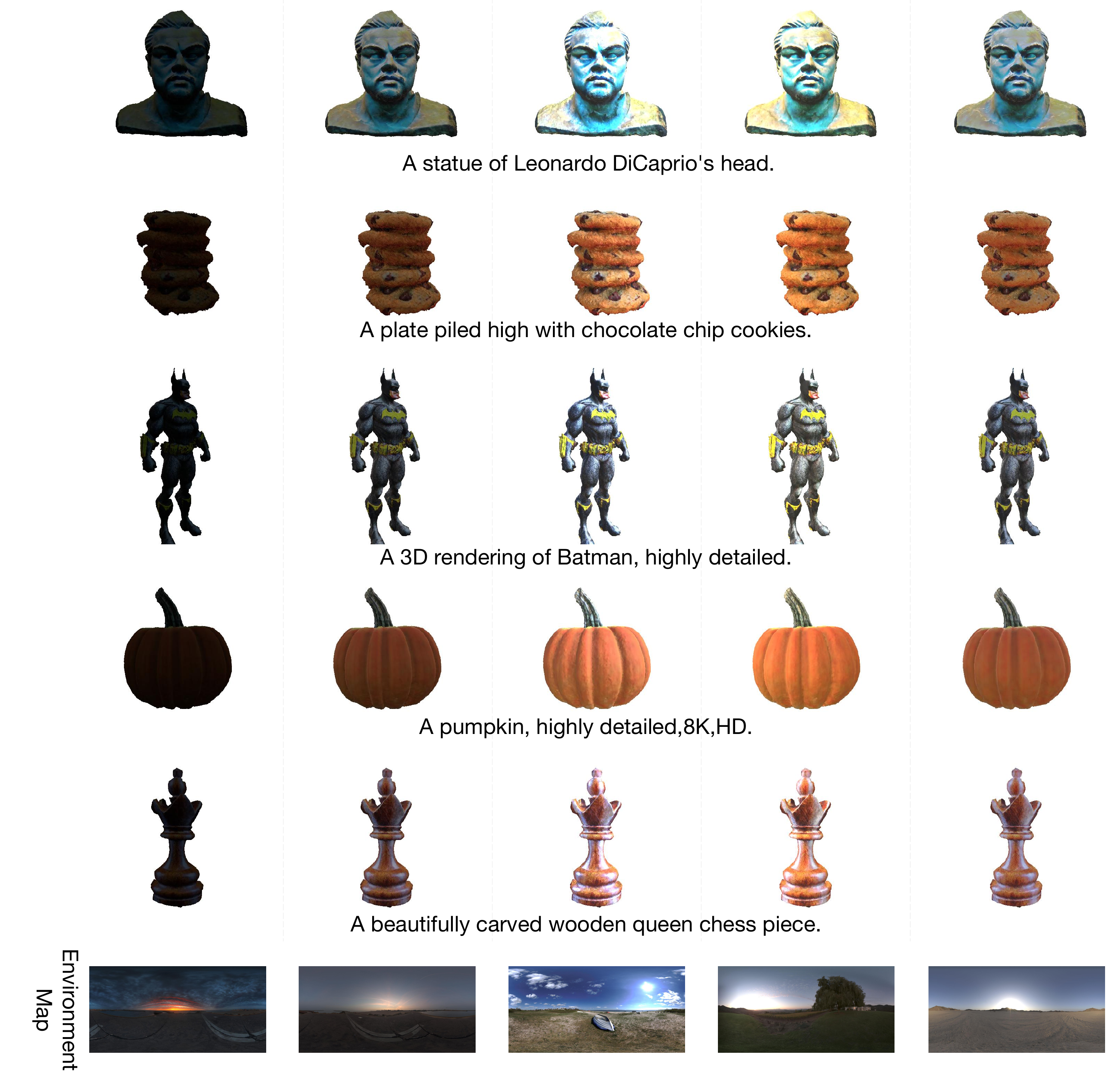}
  \caption{Rendered images of generated 3D assets under different environment maps.}
  \label{fig:env1}
\end{figure*}

\begin{figure}
  \centering
  \includegraphics[width=0.4\columnwidth]{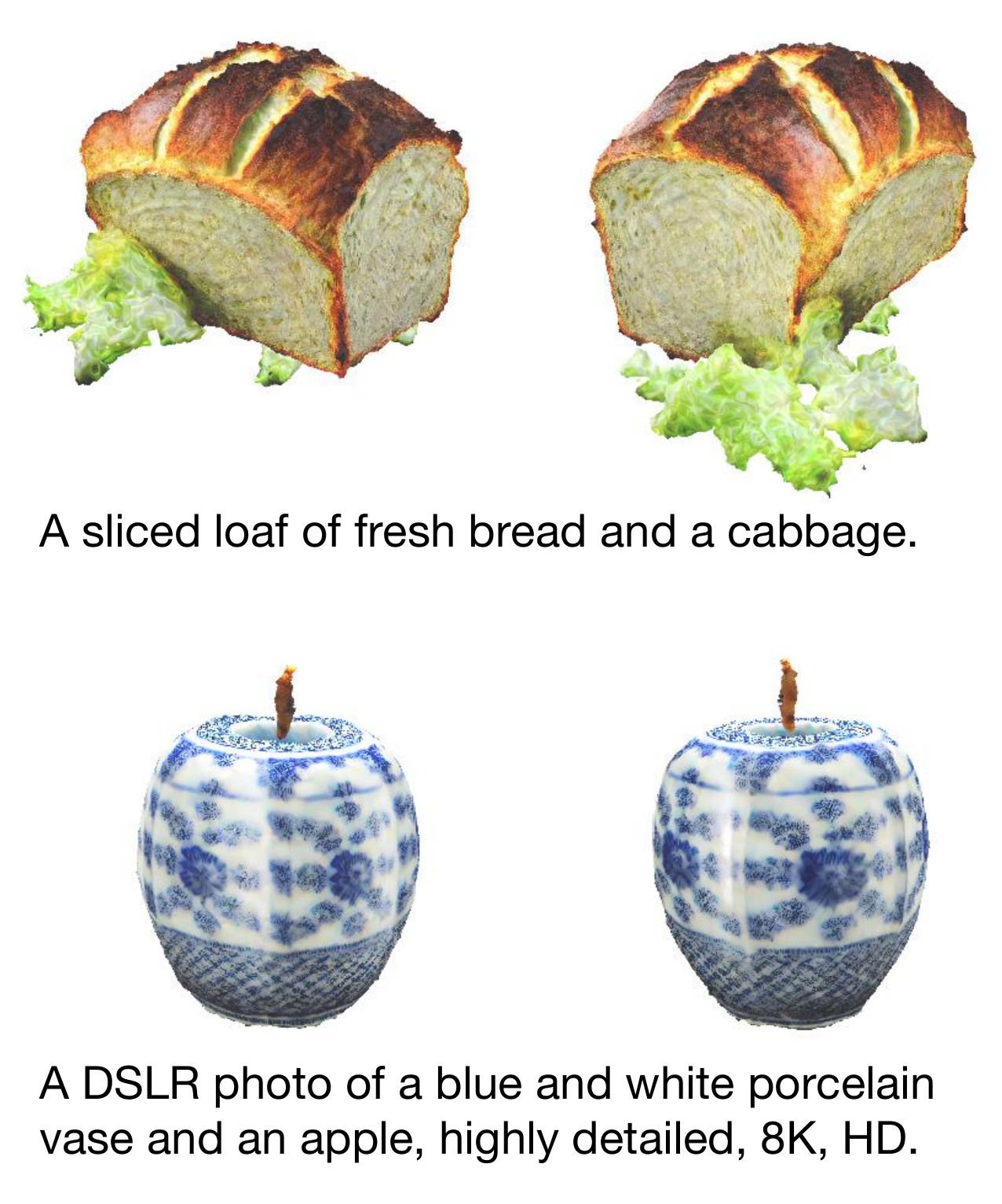}
  \caption{Bad cases of \modelname.}
  \label{fig:limitation}
\end{figure}

In this section, we present visualizations of the attention maps from various attention layers during the geometry learning stage and the appearance learning stage.
As illustrated in \figref{fig:att_layers_geo} and \figref{fig:att_layers_app}, the attention maps generated by different attention layers consistently exhibit the ability to accurately locate the shape of the foreground object. This demonstrates the robustness and effectiveness of the attention mechanism in capturing the essential features of the foreground object.
Furthermore, we observe that attention maps from different layers seem to exhibit variations in focus, with some emphasizing the foreground, some highlighting the edges, and some concentrating on the interior of the object. This divergence in focus can be attributed to the distinct roles played by different layers in the stable diffusion training process. Each layer contributes its unique perspective and emphasis, leading to variations in attention focus.
Nevertheless, despite these variations, all attention maps generated by X-Dreamer successfully locate the foreground objects with precision. This outcome provides strong evidence that our proposed module effectively guides the model's attention towards the foreground objects. By incorporating the attention mechanism into the training process, X-Dreamer ensures that the model focuses on the most relevant regions, resulting in accurate and reliable detection of the foreground objects throughout the geometry and appearance learning stages.

\subsection{Rendered Images under Different Environment Maps}

Our method, while employing a fixed environment for training the model, offers significant advantages and flexibility in the rendering of 3D assets generated by X-Dreamer. It is worth noting that these assets are not limited to a single environment; instead, they possess the capability to be rendered in diverse environments, expanding their applicability and adaptability.
The visual evidence presented in \figref{fig:env1} reinforces our claim. By utilizing different environment maps, we can readily observe the striking variations in the appearance of the rendered images of these 3D assets. This compelling demonstration showcases the inherent versatility and potential of X-Dreamer's generated assets to seamlessly integrate into various rendering engines.
The ability to effectively address the challenges posed by different lighting conditions is a crucial aspect of any rendering engine. By leveraging the capabilities of X-Dreamer, we unlock the possibility of using its 3D assets in a multitude of rendering engines. This versatility ensures that the assets can adapt and thrive in diverse lighting scenarios, enabling them to meet the demands of real-world applications in fields such as architecture, virtual reality, and computer graphics.

\subsection{Limitation}
One limitation of the proposed \modelname is its inability to generate multiple objects simultaneously. When the input text prompt contains two distinct objects, X-Dreamer may produce a mixed object that combines properties of both objects. A clearer understanding of this limitation can be gained by referring to \figref{fig:limitation}.
In the first row of \figref{fig:limitation}, our intention is for \modelname to generate ``A sliced loaf of fresh bread and a cabbage." However, the resulting output consists of a bread with a few cabbage leaves, indicating the model's difficulty in generating separate objects accurately. Similarly, in the second row of \figref{fig:limitation}, we aim to generate ``A blue and white porcelain vase and an apple." However, the output depicts a blue and white porcelain vase with a shape resembling an apple.
Nevertheless, it is important to note that this limitation does not significantly undermine the value of our work. In applications such as games and movies, each object functions as an independent entity, capable of creating complex scenes through interaction with other objects. Our primary objective is to generate high-quality 3D objects, and while the current limitation exists, it does not diminish the overall quality and utility of our approach.

\section{Conclusion}

This study introduces a groundbreaking framework called \modelname, which is designed to enhance text-to-3D synthesis by addressing the domain gap between text-to-2D and text-to-3D generation. 
To achieve this, we first propose \moduleOneShort, a module that incorporates 3D-associated information, including direction-aware text and camera parameters, into the pretrained Stable Diffusion (SD) model. 
By doing so, we enable the effective capture of information relevant to the 3D domain.
Furthermore, we design \moduleTwoShort to align the attention map generated by SD with the rendered mask of the 3D object. 
The primary objective of \moduleTwoShort is to guide the focus of the text-to-3D model towards the generation of foreground objects.
Through extensive experiments, we have thoroughly evaluated the efficacy of our proposed method, which has consistently demonstrated its ability to synthesize high-quality and photorealistic 3D content from given text prompts.

% \begin{acks}
% This work was supported by National Key R\&D Program of China (No.2023YFB4502804), the National Science Fund for Distinguished Young Scholars (No.62025603), the National Natural Science Foundation of China (No. U21B2037, No. U22B2051, No. 62072389), the National Natural Science Fund for Young Scholars of China (No. 62302411), China Postdoctoral Science Foundation (No. 2023M732948), the Natural Science Foundation of Fujian Province of China (No.2021J01002,  No.2022J06001), and partially sponsored by CCF-NetEase ThunderFire Innovation Research Funding (NO. CCF-Netease 202301).
% \end{acks}

\bibliographystyle{ACM-Reference-Format}
\bibliography{sample-base}

\end{document}